\documentclass[lettersize,journal]{IEEEtran}

\usepackage{amsmath,amsfonts}
\usepackage{algorithmic}
\usepackage{array}
\usepackage[caption=false,font=normalsize,labelfont=sf,textfont=sf]{subfig}
\usepackage{textcomp}
\usepackage{stfloats}
\usepackage{verbatim}

\usepackage{xcolor}
\usepackage{color}
\usepackage{soul}

\usepackage{graphicx}
\usepackage{svg}
\usepackage{dirtree}

\usepackage{float}
\usepackage{hyperref}
\usepackage{url}
\usepackage[capitalise]{cleveref}
\usepackage{fancyhdr}
\usepackage{lastpage}
\usepackage{booktabs,caption}
\usepackage{listings}
\usepackage{enumitem}
\usepackage{csquotes}

\usepackage{amssymb}
\usepackage{pifont} 

\usepackage{tablefootnote} 
\usepackage{multirow}
\usepackage{bm,array}
\usepackage{longtable}

\usepackage{romannum}
\AtBeginDocument{\pagenumbering{arabic}}

\definecolor{codegreen}{rgb}{0,0.6,0}
\definecolor{codegray}{rgb}{0.5,0.5,0.5}
\definecolor{codepurple}{rgb}{0.58,0,0.82}
\definecolor{backcolour}{rgb}{0.95,0.95,0.92}

\newcommand*\greencheck{\textcolor{codegreen}{\ding{52}}}
\newcommand*\redcross{\textcolor{red}{\ding{55}}}

\newcommand{\rot}[1]{\parbox[t]{2mm}{\rotatebox[origin=c]{90}{#1}}}

\definecolor{orgred}{rgb}{0.8078,0.4471,0.2314}
\definecolor{darkgreen}{rgb}{0.4157,0.6,0.333}
\definecolor{darkblue}{rgb}{0.0,0.0,0.6}
\definecolor{cyan}{rgb}{0.0,0.6,0.6}
\definecolor{light-gray}{gray}{0.80}

\lstdefinestyle{xmlStyle}{
  basicstyle=\ttfamily\scriptsize,
  columns=fullflexible,
  showstringspaces=false,
  commentstyle=\color{darkgreen},
  numbers=left,                    
  numbersep=10pt, 
  numberstyle=\tiny,
  captionpos=b,
}
\begin{document}

\title{Understanding URDF: A Dataset and Analysis}

\author{Daniella~Tola and Peter~Corke~\IEEEmembership{Fellow,~IEEE}%
\thanks{D. Tola is with the Department of Electrical and Computer Engineering, Aarhus University, Aarhus, Denmark (e-mail: dt@ece.au.dk).}%
\thanks{P. Corke is with the Queensland University of Technology (QUT) Centre for Robotics, QUT, Brisbane, Australia (e-mail: peter.corke@qut.edu.au).}%
\thanks{This work was supported by the Innovation Foundation Denmark through the MADE FAST project.}
}

\maketitle
\thispagestyle{fancy}


\begin{abstract}
As the complexity of robot systems increases, it becomes more effective to simulate them before deployment.
To do this, a model of the robot's kinematics or dynamics is required, and the most commonly used format is the Unified Robot Description Format (URDF).
This article presents, to our knowledge, the first dataset of URDF files from various industrial and research organizations, with metadata describing each robot, its type, manufacturer, and the source of the model.
The dataset contains 322 URDF files of which 195 are unique robot models, meaning the excess URDFs are either of a robot that is multiply defined across sources or URDF variants of the same robot.
We analyze the files in the dataset, where we, among other things, provide information on how they were generated, which mesh file types are most commonly used, and compare models of multiply defined robots.
The intention of this article is to build a foundation of knowledge on URDF and how it is used based on publicly available URDF files.
Publishing the dataset, analysis, and the scripts and tools used enables others using, researching or developing URDFs to easily access this data and use it in their own work.
\end{abstract}

\begin{IEEEkeywords}
robots, visualization, simulation, modeling, guidelines
\end{IEEEkeywords}


\section{Introduction}

\IEEEPARstart{M}{odeling} and simulation are key parts of the process of developing robotic systems~\cite{Sannemann&2020,Afzal&2021}. 
Their use has been increasing over the past years, as has their complexity~\cite{HeeSun&21}.
Modeling and simulation reduces the cost and risk by allowing experimentation with parameters, algorithms, and different environments before committing to expensive physical hardware.
This need has driven the development of numerous simulation tools, such as Gazebo, Webots, Unity, and RoboDK.
Each of these tools has its own native model type, (see \cref{tab:simulators_model_support}), and exchanging models between these tools can be cumbersome if a common format is not used.

The Unified Robot Description Format (URDF) was introduced in 2009 by the Robot Operating System (ROS) developers as a format to describe the kinematics, dynamics, and geometries of robots, independently of software programs~\cite{Quigley&2015}.
A URDF file is an XML-based file with an extension of \emph{.urdf}.

\begin{figure}[!t]
    \centering
    \includegraphics[width=\columnwidth]{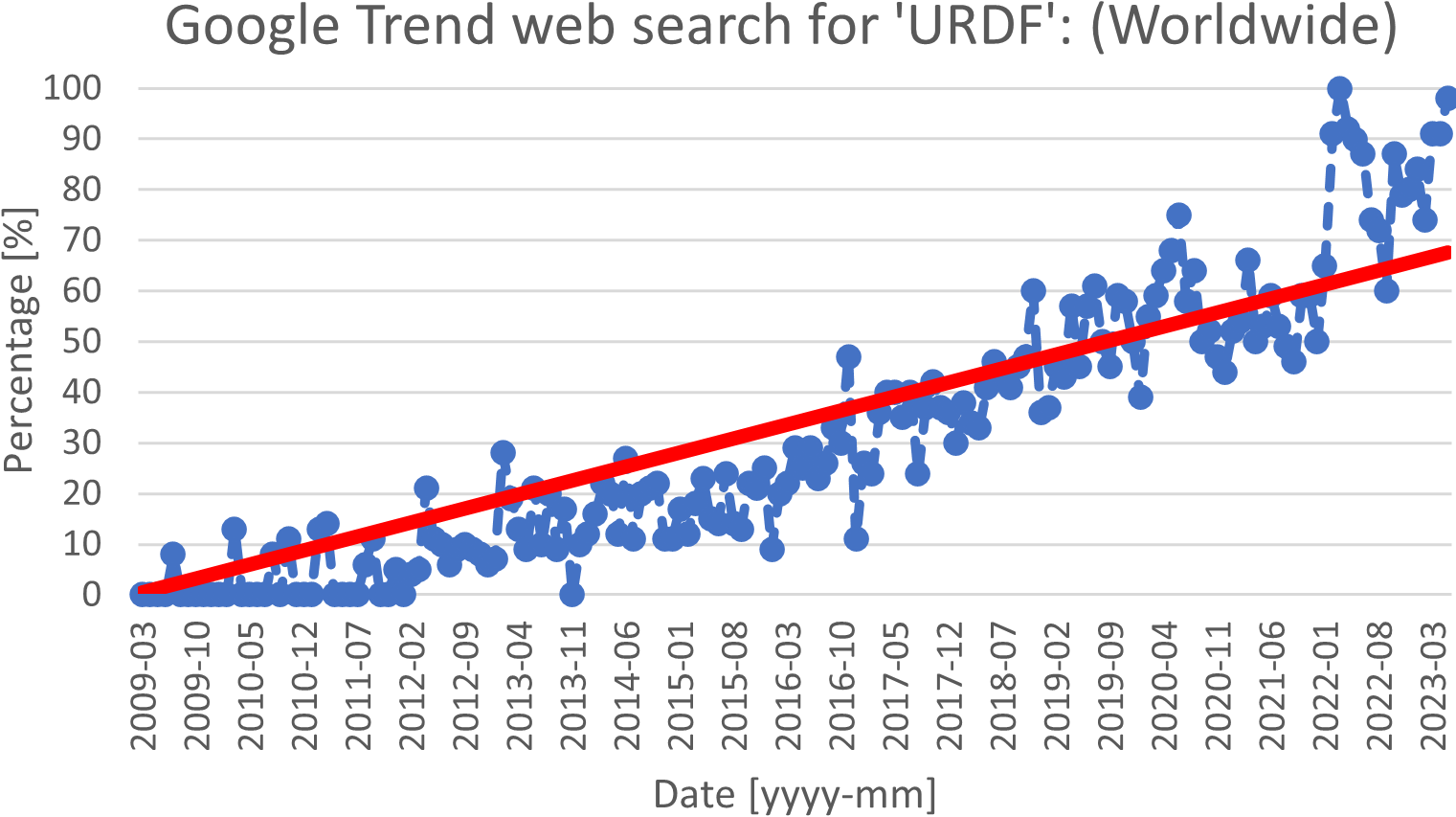}
    \caption{Google trend (\texttt{trends.google.com/trends}) for the search term `URDF' since 2009. Note that the data is an approximation of the number of Google searches. The data points at zero percent are due to insufficient data.}
    \label{fig:trend_urdf_oct2009}
\end{figure}

URDF files allow robotics developers to describe a robot using a universal format, which can be imported and exported by different tools to visualize or simulate the robot.
The number of tools supporting URDF files is growing, e.g., Unity lately (2019) added support for URDF.
\Cref{tab:simulators_model_support} shows an overview of commonly used robot simulation tools~\cite{Symeonidis2022}, their native model format, and whether or not they support URDF files, of which 11/12 tools do.
Additionally, 4/12 tools have URDF as their native model format, showing the wide adoption of URDF.
Furthermore, the trend in Google queries for the term `URDF' in \cref{fig:trend_urdf_oct2009} shows that the interest in URDF has increased over the years.

\begin{table}[!htbp]
\centering
\caption{Simulators, model format, and URDF support (as of 06/2023). Bold text represents the model formats where URDF is the native format.}
\label{tab:simulators_model_support}
\begin{tabular}{lp{2cm}p{1.2cm}}
\textbf{Simulation tool} & \textbf{Native model format} & \textbf{URDF support} \\ \hline
Gazebo  &  .sdf &  \greencheck \\ \hline
Webots  & .proto   & \greencheck   \\ \hline
Unity  & .fbx  & \greencheck \\ \hline
CoppeliaSim  & .ttm  & \greencheck   \\ \hline
RoboDK & .robot & \redcross  \\ \hline
RViz & \textbf{.urdf} & \greencheck \\ \hline
PyBullet & \textbf{.urdf}, .sdf, .mjcf & \greencheck \\ \hline
MuJoCo & .mjcf & \greencheck \\ \hline
Isaac Sim & .usd & \greencheck \\ \hline
Drake & \textbf{.urdf}, .sdf, .mjcf & \greencheck \\ \hline
MATLAB & .mat & \greencheck \\ \hline
Robotics Toolbox for Python & \textbf{.urdf} & \greencheck\\ \hline
\end{tabular}
\end{table}

There are currently limited guidelines on working with or developing URDFs~\cite{tola2023surveyURDF}.
We present our dataset and analyses to build a foundation of knowledge and practice in this area.

The contribution of this article is a novel dataset of URDF files and analyses of them.
Features of the dataset include:
\begin{itemize}
    \item Diverse collection including various types of robotic devices, such as manipulator arms, end effectors, quadrupeds, and wheeled mobile bases.
    \item Diverse sources of URDF files, including professional research and industrial organizations.
    \item Easy to use with human and machine-readable metadata describing the type of robot, manufacturer, and a URL pointing to the original URDF location.
    \item Accompanied by a Python tool and scripts for analyzing URDF files, allowing users to easily analyze new URDF files and reproduce the results in this article.
    \item Can be used to identify patterns in, and issues with the collected URDF files.
    \item Can be used for benchmarking and comparing URDF-related tools.
\end{itemize}
As there are currently no official guidelines on the naming, structure, or creation of URDF files, our analysis can be used to find the main commonalities between the URDF files and provide this guidance.
\Cref{sec:ds_urdf_file} provides a brief introduction to URDF files.
The dataset is presented in \cref{sec:ds_dataset}, and the results of various analyses of the dataset in \cref{sec:ds_analysis}.
The construction of the dataset, how to reproduce its results, and an introduction to a work-in-progress tool for analyzing URDFs, are all presented in \cref{sec:construction}.
We conclude in \cref{sec:ds_conclusion} with our main findings.


\section{What is URDF?} \label{sec:ds_urdf_file}

\noindent A URDF model is a human-readable XML file describing the kinematic structure, dynamic parameters, visual representation, and collision geometries of a robot.
The file may include references to other files that contain 3D geometries of the robot's components.
We refer to such a set of files as a URDF Bundle, introduced below in \cref{subsec:urdf_urdf_bundle}.
The main concepts of a URDF file and its associated components are described in this section.

\subsection{URDF File} \label{subsec:urdf_urdf_file}

URDF was developed to be a self-contained specification that includes all relevant modeling details of a robot within a single file.
The use of XML allows tools to easily implement support for the format, using mature XML parsers included in most programming languages.
URDF was initially introduced with ROS, but its standalone characteristic has allowed it to be adopted by many different tools, within and outside of the ROS ecosystem.
We provide a brief overview of URDF files, their syntax, and how they can be used to model a robot.
For more information, refer to the ROS wiki page on URDF\footnote{\url{wiki.ros.org/urdf/XML/model}}.

\begin{figure}[!t]
    \centering
    \includegraphics[width=0.55\columnwidth]{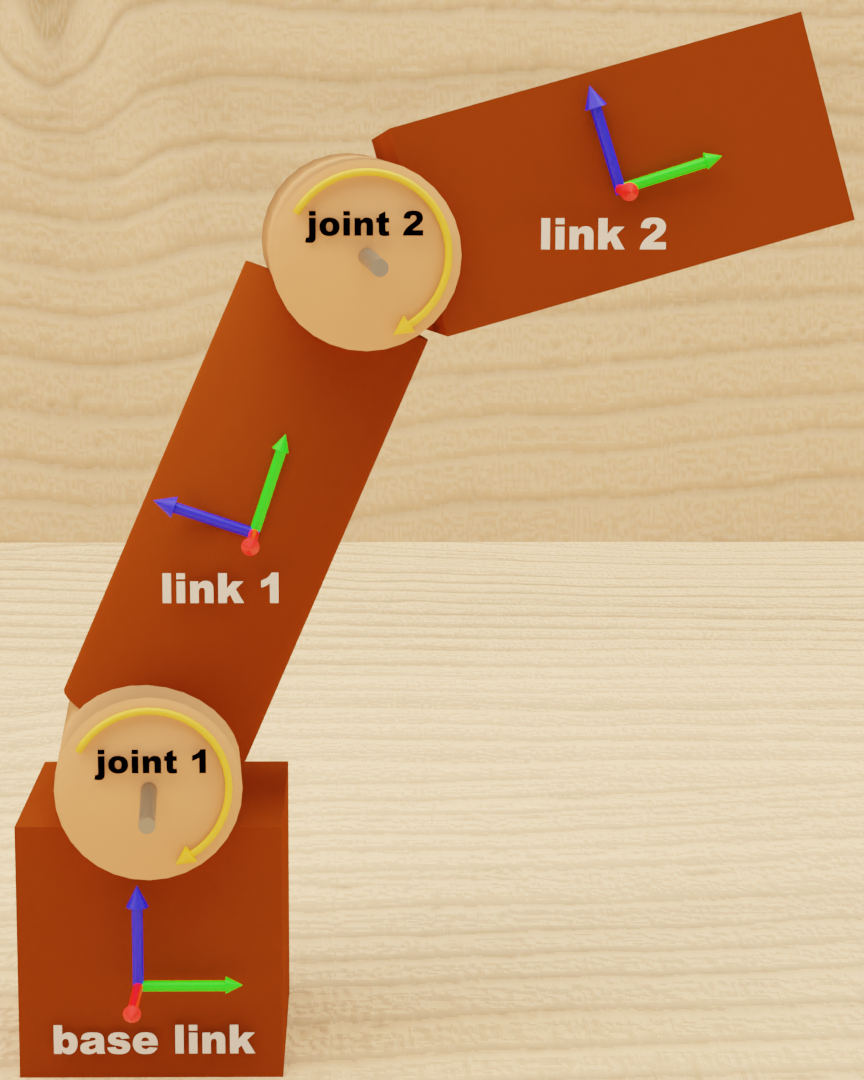}
    \caption{Visualization of the 2 DoF robot defined in Listing \ref{lst:urdf2dof}.}
    \label{fig:2dof_robot_urdf}
\end{figure}

The minimal requirements to create a URDF file are the name of the robot and a link.
To illustrate, we will consider an example of a URDF file partially shown in Listing \ref{lst:urdf2dof}.
It represents a 2 DoF planar robot, visualized using simple geometric shapes: boxes and cylinders, illustrated in \cref{fig:2dof_robot_urdf}.
The example robot has 3 links and 2 joints.

\definecolor{baselinkcolor}{rgb}{0.96,0.88,0.76}
\definecolor{link1color}{rgb}{0.93,0.93,0.95}
\definecolor{link2color}{rgb}{0.86,0.93,1}
\definecolor{joint1color}{rgb}{0.96,0.93,0.89}
\definecolor{joint2color}{rgb}{0.8,0.85,0.97}
\definecolor{others}{rgb}{0.76,0.74,0.82}
\newcommand{\coloropacity}{!65}%

\newcommand{\Hilight}[1]{\makebox[0pt][l]{\color{#1}\rule[-4pt]{0.95\columnwidth}{10pt}}} 
 \begin{minipage}{\columnwidth} 
\begin{lstlisting}[caption={URDF file contents of a 2 DoF planar robot.},label={lst:urdf2dof},style=xmlStyle]
|\Hilight{others\coloropacity}|<?xml version="1.0" encoding="utf-8"?>
|\Hilight{others\coloropacity}|<robot name="2 DOF planar robot">
|\Hilight{baselinkcolor\coloropacity}| <link name="base link">
|\Hilight{baselinkcolor\coloropacity}\hspace{0.2cm}|  <visual>
|\Hilight{baselinkcolor\coloropacity}\hspace{0.4cm}|   <origin xyz="0 0 0.25"/>
|\Hilight{baselinkcolor\coloropacity}\hspace{0.4cm}|   <geometry>
|\Hilight{baselinkcolor\coloropacity}\hspace{0.65cm}|    <box size="0.5 0.5 0.5"/>
|\Hilight{baselinkcolor\coloropacity}\hspace{0.4cm}|   </geometry>
|\Hilight{baselinkcolor\coloropacity}\hspace{0.2cm}|  </visual>
|\Hilight{baselinkcolor\coloropacity}| </link>
...
|\Hilight{joint1color\coloropacity}| <joint name="joint 1" type="continuous">
|\Hilight{joint1color\coloropacity}\hspace{0.2cm}|  <parent link="base link" />
|\Hilight{joint1color\coloropacity}\hspace{0.2cm}|  <child link="link 1" />
|\Hilight{joint1color\coloropacity}\hspace{0.2cm}|  <axis xyz="0 1 0" />
|\Hilight{joint1color\coloropacity}\hspace{0.2cm}|  <origin xyz="0 0 0.5"/>
|\Hilight{joint1color\coloropacity}| </joint>
...
|\Hilight{others\coloropacity}|</robot>
\end{lstlisting}
\end{minipage}

\subsubsection{Links}
are rigid bodies that can be connected using joints, and are described by inertial, visual, and collision properties\footnote{\url{wiki.ros.org/urdf/XML/link}}.
The inertial properties describe the mass of the link, the center of mass position, and the moments and products of inertia.
The visual and collision properties are described in more detail in \cref{subsec:urdf_geom_meshes}.
URDF links can only be represented using rigid bodies and not deformable ones.

The names of the links of the example robot are ``\texttt{base link}'', ``\texttt{link 1}'', and ``\texttt{link 2}''.
We look at the ``\texttt{base link}'' to exemplify how a link is specified, see Listing \ref{lst:urdf2dof} at lines 3-10.
This link represents the fixed base of the robot, where its visual properties are defined by an \texttt{origin} and a \texttt{geometry} consisting of a \texttt{box}.
The size of the \texttt{box} is specified by values of its three side lengths.
The only required specification of a link is its name, which \textit{must} be unique. 

\subsubsection{Joints}
connect two links, a parent and child link.
The parent is the link closer to the base (or root) of the mechanism, the child is closer to the tool tip.
The main specifications of joints are the type (kinematics), dynamics, and safety limits\footnote{\url{wiki.ros.org/urdf/XML/joint}}.
Supported joint types are: revolute, continuous, prismatic, fixed, floating, and planar.

The names of the example robot's joints are ``\texttt{joint 1}'' (see Listing \ref{lst:urdf2dof} at lines 12-17) and ``\texttt{joint 2}''.
The joints are continuous, meaning they are revolute joints with no motion limits.
The axis property specifies the joint axis, which in this example is a rotation about the y-axis.
The required specifications of a joint are its name, type, and names of the parent and child links.

\subsection{Visual and Collision Geometries} \label{subsec:urdf_geom_meshes}

Geometric objects are used to represent the shape of a robot's links for visualization or collision purposes, and are called meshes.
They comprise a set of polygons (typically triangles) that form the surface of the object.
The more polygons in a mesh, the higher the level of detail of the shape, but at the expense of rendering and calculation time.

Meshes can be provided using different Computer-aided design (CAD) file types.
Each file type has a different internal format, with its own benefits and limitations, and should therefore be chosen depending on the application of use.
A commonly used format for both visualization and collision meshes in URDF is STL (with file extension \textit{.stl}), which represents 3D surface geometries using only triangles and no color or texture information.
Another format, COLLADA (with file extension \emph{.dae}), is typically used for visualization as it supports both color and texture information.
The OBJ format (with file extension \emph{.obj}), supports color, texture, and free-form curves, allowing for higher levels of detail for visualization, however, the color and texture data is stored in a separate \emph{(.mtl)} file.

In some applications of URDF, collision detection is required, while in others the URDF model is solely used for visualization purposes.
Depending on the application, different types of meshes are included in the URDF Bundle.
For example, it is common to use both STL and COLLADA meshes, as the STL meshes reduce the computation and rendering time, while representing the approximate form of the shape in the collision calculations, and the COLLADA meshes can at the same time provide high-quality visualizations of the robot. 

\subsection{URDF Bundle} \label{subsec:urdf_urdf_bundle}
\begin{figure}[!t]
    \centering
    \includegraphics[width=\columnwidth]{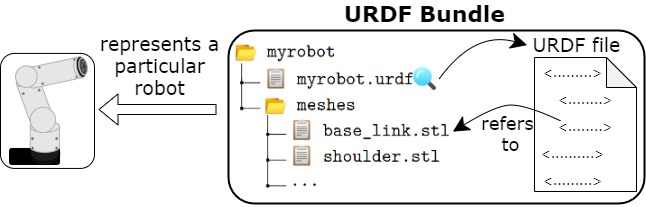}
    \caption{Example of a URDF Bundle. It is common for mesh files to be located in a separate folder.}
    \label{fig:urdf_bundle}
\end{figure}

A URDF robot model can consist of the URDF file and mesh files describing the physical appearance of the robot's links.
We distinguish between the URDF file itself (with the \emph{.urdf} file extension) and the set of files comprising the URDF file and meshes, by referring to the latter as a \emph{URDF Bundle}, see \cref{fig:urdf_bundle}.
This URDF Bundle contains the URDF file called \emph{myrobot.urdf} together with the meshes of the robot located within the \emph{meshes} folder.
The URDF file refers to the geometric mesh files of the different links using relative paths.

\subsection{Xacro}
Xacro\footnote{\url{wiki.ros.org/xacro}} is a macro language for XML commonly used in ROS, which allows constructing URDF Bundles.
Some of the key operations that xacro can perform are variable substitution, math calculations, file inclusion, and conditional blocks.
Xacro provides tags that can be used to configure URDF files, based on the application, to reduce redundancy and simplify the maintainability of the models.
These tags are defined in a xacro-based URDF file using the xacro extension \emph{(.xacro)}.
The xacro preprocessor is an executable used to generate URDF Bundles by interpreting the xacro tags and using input values.
The preprocessor allows combining multiple xacro files, which is especially beneficial when dealing with complex robotic structures~\cite{Albergo&2022}.


\section{Dataset} \label{sec:ds_dataset}

\noindent The URDF Bundles in this dataset consist solely of robots with 3D geometric meshes representing either the visual properties, or both the visual and collision properties of the robots.
The dataset is publicly available in the GitHub repository\footnote{\url{github.com/Daniella1/urdf_files_dataset}}.

\subsection{Definitions}

\begin{description}
    \item[URDF Bundle vs. robot:] Any particular robot can be described by a URDF Bundle. 

    \item[URDF variant:] URDF files can represent different features of the same robot, depending on the application of the URDF.
    For example, the Kuka LBR Iiwa 14 robot has URDF variants such as \textit{spheres collision}, \textit{no collision}, \textit{spheres dense elbow collision}, etc., and the Atlas robot has URDF variants of \textit{convex hull} and \textit{minimal contact}. 
    These URDF variants seem to be created for different applications or simulation purposes.
    As not all of these URDF variants include explanations, it may be difficult in some cases to choose the most appropriate URDF file for a given application.

    \item[Multiply defined robot:] Some sources provide URDF Bundles representing the same robot.
    For example, both sources \texttt{matlab} and \texttt{ros-industrial} provide the Universal Robots UR5e robot.
    We characterize such a robot as a multiply defined robot.
    The files of a multiply defined robot are not necessarily identical.
    
\end{description}

\subsection{Overview}

\begin{figure}[!t]
    \centering
    \includegraphics[width=\columnwidth]{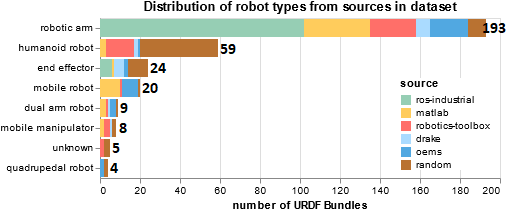}
    \caption{Number of URDF Bundles of each robot type.}
    \label{fig:robot_types}
\end{figure}

The URDF Bundles in the dataset have been gathered from six different sources described below.
\Cref{tab:urdf_dataset_overview} shows the total number of URDF Bundles and their URDF variants from each source, and \cref{fig:robot_types} shows the number of URDF Bundles by robot type.

\begin{table}[!htbp]
\centering
\caption{An overview of the URDF Bundles (incl. variants) and URDF variants in the dataset.}
\label{tab:urdf_dataset_overview}
\begin{tabular}{l|c|c}
\textbf{Source} & \textbf{\#URDF Bundles} & \textbf{\#URDF Variants} \\ \hline
\texttt{ros-industrial}  &  108 & 1 \\ 
\texttt{matlab} & 52 &2 \\ 
\texttt{robotics-toolbox} & 44 & 15 \\ 
\texttt{drake} & 16 & 12\\ 
\texttt{oems}  &  35  & 6 \\ 
\texttt{random}  &  67 &  39 \\ \hline
\textbf{Total} & \textbf{322} & \textbf{75} \\\hline
\end{tabular}
\end{table}

\begin{description}
    \item[ros-industrial:] is an open-source project with the goal of supporting ROS for manufacturing and automation\footnote{\url{github.com/ros-industrial}}.
    The project builds on ROS, and includes URDF Bundles and drivers for specific robots. 
    The developers and contributors of the project are mostly research organizations.

    \item[matlab:] is a commercial platform for programming and numeric computing which contains a number of toolboxes\footnote{\url{mathworks.com/help/robotics/ref/importrobot.html}}. The Robotics System Toolbox is shipped with URDF Bundles of commonly-used robots.
    These Bundles are also part of the public dataset and have their own individual licensing.
    The MATLAB version used in the construction of this dataset is R2022b.
    
    \item[robotics-toolbox:] is an open-source Python toolbox providing various tools for working with kinematics and dynamics, visualizations, and path planning~\cite{Corke&21}. The toolbox is developed and maintained by researchers.

    \item[drake:] is an open-source Python and C++ toolbox providing tools for modeling dynamical systems, working with kinematics and dynamics, and solving mathematical programs\footnote{\url{github.com/RobotLocomotion/drake}}. The toolbox is developed and maintained by researchers and developers.
    The URDF Bundles provided by \texttt{drake} are modified from other sources, where they have appended a readme or license file to describe the modifications and origins of the files.

    \item[oems:] is an assortment of URDF Bundles provided directly by the Original Equipment Manufacturers (OEMs) of the robots.
    
    \item[random:] is an assortment of URDF Bundles from various GitHub repositories. These URDF Bundles may have been developed by researchers, developers, or others. 

\end{description}

\subsection{Structure}

\begin{figure}[!t]
    \centering
    \includegraphics[width=0.7\columnwidth]{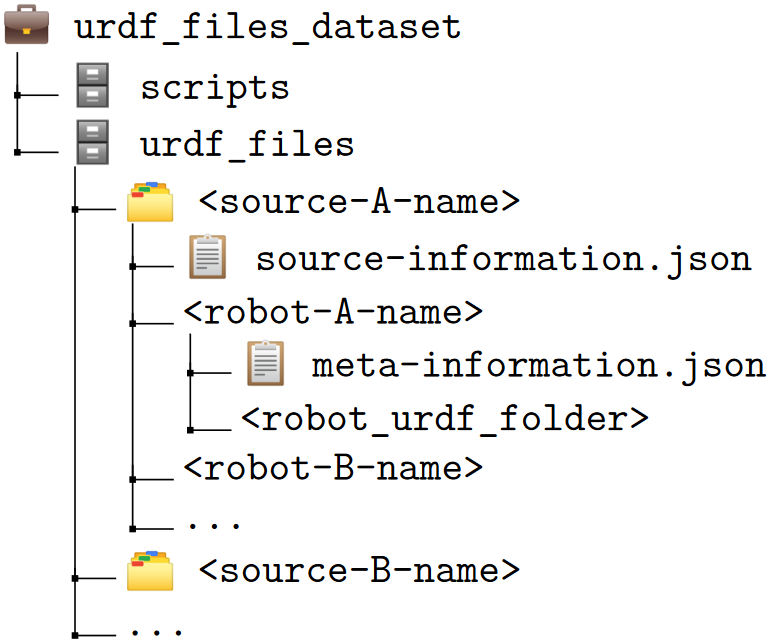}
    \caption{Structure of the \emph{urdf\_files} directory.}
    \label{fig:dataset_urdf_files_dir_tree}
\end{figure}

The dataset consists of two subdirectories: \emph{scripts} and \emph{urdf\_files}.
The \emph{scripts} directory contains a number of Python scripts used to analyze the dataset and generate the results presented in this article.
The \emph{urdf\_files} directory contains all of the URDF Bundles in the dataset categorized by the source, as shown in \cref{fig:dataset_urdf_files_dir_tree}.
Information about the source and robots (URDF Bundles) is included using human and machine-readable JSON files.
The source information (name and URL) is described using a \texttt{source-information.json} file.
Each source contains subdirectories with URDF Bundles, described using \texttt{meta-information.json} files.
The meta information includes the name, type, manufacturer, relative URDF file location, ID, source URL, and whether the URDF Bundle has manually been generated using xacro while creating the dataset.

\section{Analysis} \label{sec:ds_analysis}
\subsection{Manufacturers}

\begin{figure}[!t]
    \centering
    \includegraphics[width=\columnwidth]{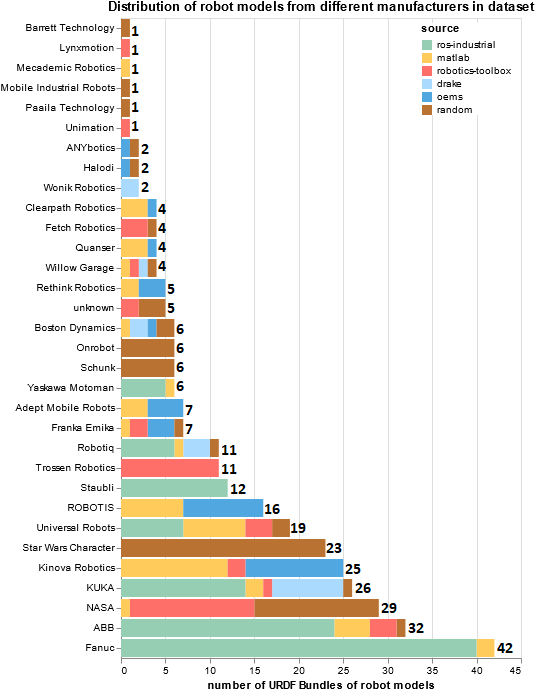}
    \caption{Bar plot of manufacturers and the number of URDF Bundles of their robot models represented in the dataset.}
    \label{fig:manufacturers_information}
\end{figure}

In total there are 32 different manufacturers of the robots in the dataset, see \cref{fig:manufacturers_information} for the distribution.
Of these:
\begin{itemize}
    \item One is unknown, as the robot name is not provided, and four are test URDF files that do not represent a robot. These are all marked as \textit{Unknown} in the dataset.
    \item One is fictional, as the robot is R2-D2 from the movie Star Wars, marked as \textit{Star Wars Character} in the dataset.
    \item Three are no longer operational (at the time of writing); these are \textit{Unimation}, \textit{Willow Garage}, and \textit{Adept Mobile Robots}.
\end{itemize}

We have divided the manufacturers into four application categories based on what their robots are mainly used for:
\begin{description}

    \item[industry:] the robots are mainly used in applications for making profits. Examples are robots in manufacturing and service robots.
    
    \item[non-profit:] the robots are mainly used for research, education, or for hobbyists. 
    
    \item[industry \& non-profit:] the robots are widely used for both industrial and non-profit purposes.
    
    \item[other:] represents manufacturers that do not fit into any of the previous categories. This category is not taken into account when calculating general numbers on manufacturers. The two manufacturers in this category are \textit{Unknown} and \textit{Star Wars Character}.
    
\end{description}
This categorization is based on our subjective opinion, and considers the following information:
\begin{itemize}
    \item using our own knowledge about the manufacturers,
    \item checking each manufacturer's website to see the type of customers they are targeting. For example, some of the manufacturers explicitly use the keywords \textit{industrial robots} or \textit{research} or \textit{education},
    \item looking at the types of robots they develop. For example, if they develop small robot kits, they can be associated with research or hobbyists.
\end{itemize}

\Cref{tab:overview_provide_urdf_main_usage_manufacturer} shows the number and percentage of manufacturers that provide a URDF Bundle for at least one of their robots.
40\% of the industry-targeted manufacturers supply at least one URDF Bundle (77\% of the dataset).
86\% of manufacturers targeting non-profit applications supply a URDF Bundle of at least one of their robots (12\% of the dataset).
The list of manufacturers, their categorization, and the procedure to determine if a manufacturer supplies URDF Bundles, can be found in the GitHub repository\footnote{\url{github.com/Daniella1/urdf_dataset_results_material}}.
It is important to note that these numbers may differ in reality, as some manufacturers provide URDF files through collaborations with third party organizations.

\newcolumntype{D}{>{\centering\arraybackslash}p{8em}}
\begin{table}[!htbp]
\centering
\caption{Number of manufacturers that supply a URDF Bundle for at least one of their robots, based on the target application of the manufacturer's robots. The percentage of robots describes how many robots in the dataset represent models from manufacturers of the specific target application.}
\label{tab:overview_provide_urdf_main_usage_manufacturer}
\begin{tabular}{l|D|D}
\textbf{Application} & \textbf{URDF supplied by manufacturer} & \textbf{\% of robots from target application}\\ \hline
industry & 8/20 & 77 \\
non-profit & 6/7 & 12  \\
industry \& non-profit & 2/3 & 10  \\ 
other & ---  & 1  \\ \hline
\textbf{Total} & 16/30 & 100  \\ \hline
\end{tabular}
\end{table}

\newcolumntype{C}{>{\centering\arraybackslash}p{4.2em}}
\begin{table}[!htbp]
\centering
\caption{Number of manufacturers that provide URDF Bundles versus how many manufacturers link to these URDF Bundles directly from their website. As the table shows only 4/16 manufacturers provide information about URDF on their websites. The column `no website' shows the manufacturers that are no longer operational (at the time of writing).}
\label{tab:manufacturer_provide_urdf_and_search}
\begin{tabular}{p{1.7cm}r|CcCC} 
\multirow{2}{*}{\textbf{Provides URDF}} & \multirow{2}{*}{\textit{total}} &
\multicolumn{4}{c}{\textbf{`URDF' in search}}  \\ 
\cline{3-6}
&& \textbf{not found} & \textbf{found} & \textbf{no search bar} & \textbf{no website} \\ \hline
\textbf{yes} & 16 & 7 & 4 & 3 & 2 \\ 
\textbf{no} & 14 & 9 & 0 & 4 & 1 \\ \hline
\end{tabular}
\end{table}

Given the value of simulation we would expect manufacturers to help users by making URDF Bundles readily available and also provide references or relevant information on URDF on their website.
To test this hypothesis we went through all the manufacturers from the dataset and gathered information on this, where we checked if the term `URDF' could be found when searching on their websites. 
The results are shown in \cref{tab:manufacturer_provide_urdf_and_search}.

\subsection{Common URDF Folder Structures}

\begin{figure}[!t]
    \centering
    \includegraphics[width=0.55\columnwidth]{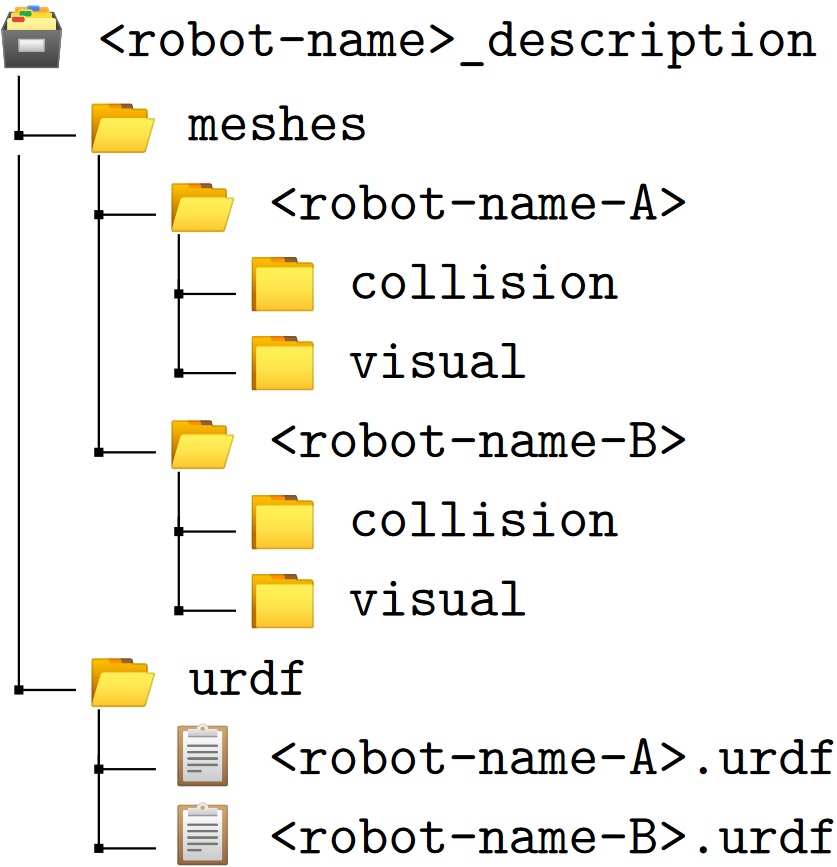}
    \caption{Folder structure A for multiple URDF Bundles.}
    \label{fig:folder_structure_A_multiple_robots}
\end{figure}

Four of the most commonly used folder structures in the dataset were identified and quantified, see \cref{tab:folder_structure_sources}.
Folder structure A, shown in \cref{fig:folder_structure_A_multiple_robots}, characterizes the structure used for multiple URDF Bundles.
The same structure can be used for single URDF Bundles, where there is only one \emph{.urdf} file in the \texttt{urdf} folder, and the \texttt{collision} and \texttt{visual} folders are directly under the \texttt{meshes} folder.
In some cases, the \texttt{collision} and \texttt{visual} folders do not exist, and the CAD files are placed directly in the \texttt{meshes} folder.
Multiple URDF Bundles within a directory, typically represent a particular robot and its different URDF variants.
Structure B is similar to structure A but the name of the root folder is \texttt{<manufacturer-name>\_<robot-name>\_support}.
This structure is also used for both single or multiple URDF Bundles.
Structure C is nearly identical to structure A, however, the \texttt{urdf} folder is instead named \texttt{robots}.
Structure D is also similar to structure A, however, the root directory name ends with \texttt{visualization} instead of \texttt{description}.
Structures C and D are only used for single URDF Bundles.
Structure D was found to only be used for end effectors.
It is important to note that not all the URDF folders follow the described structures, we have chosen to only present the most commonly used structures and count their occurrence in the dataset.

\begin{table}[!htbp]
\centering
\caption{Four of the most common folder structures in sources and the number of URDF Bundles using them.}
\label{tab:folder_structure_sources}
\begin{tabular}{l|cccc}\hline
\multirow{2}{*}{\textbf{Source}} & \multicolumn{4}{c}{\textbf{Structure}}\\
\cline{2-5}
& \textbf{A} & \textbf{B} & \textbf{C} & \textbf{D} \\ \hline
\texttt{ros-industrial} & 1 & 55 & -- & 4 \\
\texttt{matlab} & 17 & 5 & 3 & --  \\
\texttt{robotics-toolbox} & 8 & -- & 1 & -- \\
\texttt{drake}  & 3 & -- & -- & -- \\
\texttt{oems}  & 9 & -- & 1 & -- \\
\texttt{random} & 4 & -- & 1 & 4 \\ \hline
\textbf{Total} & 42 & 60 & 6 & 8 \\ \hline
\end{tabular}
\end{table}

\subsection{Xacro Generated URDF Bundles}

The number of URDF Bundles in the dataset that were generated with and without xacro are shown in \cref{tab:source_xacro_information}.
As the results show most of the URDF Bundles (95\%) have been generated using xacro.

\newcolumntype{C}{>{\centering\arraybackslash}p{4em}}
\begin{table}[!t]
\centering
\caption{Number of URDF Bundles that have been generated in the dataset with and without xacro. The column `By us' represents the URDF Bundles that we generated while creating the dataset. The column `By others' represents URDF Bundles created by others.}
\label{tab:source_xacro_information}
\begin{tabular}{l|C|CC}
\multirow{2}{*}{\textbf{Source}} & \textbf{By us} & \multicolumn{2}{c}{\textbf{By others}}  \\
\cline{2-4}
& \textbf{using xacro} & \textbf{using xacro} & \textbf{without xacro}\\ \hline
\texttt{ros-industrial} & 90  & 18 & 0 \\ 
\texttt{matlab} & 0 & 49 & 3 \\ 
\texttt{robotics-toolbox} & 17 & 24 & 3 \\ 
\texttt{drake} & 0  & 16 & 0 \\
\texttt{oems} & 26 & 6 & 3 \\ 
\texttt{random} & 12 & 49 & 6 \\ \hline
\textbf{Total} & 145 & 162 & 15 \\ \hline
\end{tabular}
\end{table}

\subsection{Parsing the URDF Files} \label{subssec:analysis_parsing}

Each URDF file was validated and the errors and warnings were extracted, using the official ROS URDF parser\footnote{\url{github.com/ros/urdfdom}} from the \texttt{urdfdom} package\footnote{\url{anaconda.org/conda-forge/urdfdom}} version 3.1.0.
Information about validation using other URDF parsers is provided in \cref{sec:construction}.
The results showed that 11/322 URDF files failed the parser, with the errors summarized in \cref{fig:piechart_parsing_issues}.
The source with the highest number of 4 URDF files failing is \texttt{random}.
Only one URDF file (from \texttt{drake}) resulted in a warning, which was that the link material was undefined.
A more detailed description of the errors follows below:
\begin{description}
    \item[(A) Issue with joint limits:] revolute and prismatic joints require joint limit specification of effort and velocity.
    If none of these attributes is provided, then the URDF file results in an error.
    
    \item[(B) No link elements found in URDF file:] at least one link is required in a URDF file, otherwise there is no kinematic structure represented.

    \item[(C) Non-unique link:] each link name must be unique in a URDF file to distinguish between them when assigning them as parent or child links in joints.

    \item[(D) No name given for robot:] the name of the robot in the URDF file must be specified.

    \item[(E) Parent link not found:] a joint requires both a parent and child link for the parser to be able to place the joint in the kinematic structure.

    \item[(F) XML parsing failed:] model parsing of the xml file failed.
\end{description}

\begin{table}[!htbp]
\centering
\caption{The parsing results of the URDF files using the official ROS parser. The errors are described in \cref{subssec:analysis_parsing}.}
    \label{fig:piechart_parsing_issues}
\begin{tabular}{c|c|l}
   \textbf{Error}  &  \textbf{\#URDF files} &  \textbf{Sources} \\ \hline
   A & 3 & \texttt{drake} \\
   B & 4 & \texttt{random} (2), \texttt{robotics-toolbox} (2) \\
   C & 1 & \texttt{random} \\
   D & 1 & \texttt{oems} \\
   E & 3 & \texttt{random} \\ 
   \multirow{2}{*}{F} & \multirow{2}{*}{11} &  \texttt{random} (4), \texttt{oems} (3), \\ & & \texttt{drake} (2), \texttt{robotics-toolbox} (2) \\ \hline
\end{tabular}
\end{table}

\subsection{CAD Files and Meshes}
\begin{figure}[!t]
    \centering
    \includegraphics[width=\columnwidth]{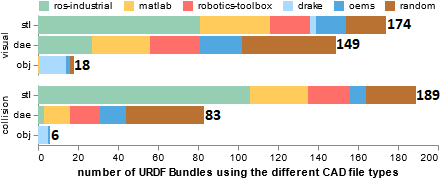}
    \caption{Number of URDF Bundles referring to different CAD file types based on mesh use.}
    \label{fig:cad_types_sources}
\end{figure}

The number of URDF Bundles with mesh files based on the CAD file type is shown in \cref{fig:cad_types_sources}.
STL is the most commonly used, followed by COLLADA and OBJ.
Not all URDF Bundles contain collision meshes, as there are 341 URDF Bundles with visual meshes, but only 278 with collision meshes.
The most commonly used CAD type for collision meshes is STL, which can be expected as collision checks do not necessarily require high precision models or colors.
Furthermore, some URDF Bundles combine different CAD file types for visual meshes, explaining the fact that there are 341 URDF Bundles with visual meshes compared to the total of 322 URDF Bundles in the dataset.

\subsection{Multiply Defined Robots} \label{subsec:analysis_duplicate_robots}
In total 60 robots have multiple definitions with 130 URDF Bundles from different sources, implying an average of 2.2 sources per multiply defined robot.
\Cref{tab:source_duplicate_n_robots} shows the number of URDF Bundles of multiply defined robots across the sources, and of them the 6 URDF Bundles with parsing errors.
4/6 URDF Bundles with issues originated from the same two multiply defined robots, meaning that the specific robots' URDF Bundles failed for two sources.

\begin{table}
\centering
\caption{Number of URDF Bundles of multiply defined robots across the sources, and the number of them that result in errors when parsing with the ROS parser. The numbers correlated between the same source indicate the number of URDF Bundles from that source that represent a multiply defined robot between that source and any other.}
\label{tab:source_duplicate_n_robots}
\begin{tabular}{r|*{7}{c}}
\textbf{Source} & \romannum{1} & \romannum{2} & \romannum{3} & \romannum{4} & \romannum{5} & \romannum{6} & \texttt{\#err}\\
\hline
\texttt{ros-industrial} (\romannum{1}) & 14 & & & & & & 0 \\
\texttt{matlab} (\romannum{2}) & 11 & 37 & & & &  & 0\\
\texttt{robotics-toolbox} (\romannum{3})  & 4 & 7 & 13 &  & & & 2 \\
\texttt{drake} (\romannum{4}) & 2 & 1 & 1 & 3 & &  &1\\
\texttt{oems} (\romannum{5}) & 0 & 22 & 2 & 0 & 25 &  &1\\
\texttt{random} (\romannum{6}) & 2 & 6 & 8 & 1 & 3 & 12 & 2 \\
\hline
\end{tabular}
\end{table}

\Cref{tab:duplicates_diff_robots} is a comparison of features of the URDF Bundles of the multiply defined robots.
It shows that the feature with the largest number of differences is the number of lines in the URDF files.
The number of joints and links differed in 9 multiply defined robots, where we found some of the differences were additional joints and links representing the world and end effectors.
One of the surprising results is that 11 multiply defined robots had different forward kinematics.

\begin{table}[!htbp]
\centering
\caption{Feature differences across the multiply defined robots. The \emph{any} feature indicates at least one difference was found between the URDF Bundles, while the \emph{any excl. lines} represents the same, however, excluding the number of lines.}
\label{tab:duplicates_diff_robots}
\begin{tabular}{l|c}
\textbf{Feature discrepancies} & \textbf{\#Robots} \\ \hline
number of joints & 9 \\ 
number of links & 9 \\ 
CAD file type & 6 \\ 
forward kinematics & 11 \\ 
number of lines & 38 \\ 
any & 38 \\ 
any excl. lines & 12 \\ \hline 
\end{tabular}
\end{table}

\subsection{Identical Files}
The Linux command \texttt{fdupes}\footnote{\url{linux.die.net/man/1/fdupes}} was used to find identical files across the different sources, see \cref{tab:fdupes_sources_and_filetypes}.
The \texttt{fdupes} command finds identical files within a given set of directories by comparing file sizes, MD5 signatures, and comparing the files byte-by-byte.
Before running the \texttt{fdupes} command on the files, we removed white spaces, tabs, and changed the carriage return of all files to be DOS (CRLF).

\begin{table}[!htbp]
\centering
\caption{Robots containing identical files (using \texttt{fdupes}) across sources.}
\label{tab:fdupes_sources_and_filetypes}
\begin{tabular}{l|cccc}
\textbf{Source}  & \textbf{.urdf}  & \textbf{.stl} & \textbf{.dae} & \textbf{.obj}\\ \hline
\texttt{ros-industrial} & 0 & 13 & 9 & 0 \\ 
\texttt{matlab}  & 0 & 19 & 9 & 0 \\ 
\texttt{robotics-toolbox} & 1 & 6 & 9 & 0 \\ 
\texttt{drake} & 0 & 1 & 0  & 0 \\ 
\texttt{oems}   & 0 & 5 & 4 & 1 \\  
\texttt{random}  & 1 & 5 & 5 & 1 \\ \hline
\end{tabular}
\end{table}

Examples of identical files across sources for two robots are shown in \cref{tab:identical_files_robots}.
As the table illustrates, the sources of the \textit{.stl} files differ from the sources of the \textit{.dae} files, and the number of identical files across the sources varies, making it difficult to trace back where these files initially originate from.

\newcolumntype{e}{>{\centering\arraybackslash}p{2em}}
\newcolumntype{f}{>{\centering\arraybackslash}p{3.5em}}
\begin{table}
\centering
\caption{Number of identical files across sets of sources for two robots. The \{\} contain the sources that share the specified number of identical files.}
\label{tab:identical_files_robots}
\begin{tabular}{c|ep{3.2cm}f}
\textbf{Robot} & \textbf{File type} & \textbf{Sources} & \textbf{\#Identical files} \\ \hline
\multirow{7}{*}{PR2}   
    & \multirow{3}{*}{stl}  & \{matlab, drake\} & 8 \\
        & & \{matlab, robotics-toolbox, drake, random\} & \multirow{2}{*}{31} \\ 
        & & \{robotics-toolbox, random\} & 9 \\ \cline{2-4}
    & \multirow{2}{*}{dae} & \{matlab, random\} & 18 \\ 
        & &  \{matlab, robotics-toolbox, random\} & \multirow{2}{*}{5} \\ \hline
\multirow{3}{*}{Franka Panda} 
    & \multirow{2}{*}{stl} & \{matlab, oems, random\} & 9 \\ 
        & & \{matlab, random\} & 1\\ \cline{2-4}
 & dae & \{robotics-toolbox, oems\} & 10 \\ \hline
\end{tabular}
\end{table}

\subsection{Model Structure}

\Cref{tab:robot_type_avg_model_info} shows the average number of links and joints of the different types of robots.
The numbers give an idea of the complexity of modeling such robots using URDF.

\begin{table}
\centering
\caption{Robot type and robot model information.}
\label{tab:robot_type_avg_model_info}
\begin{tabular}{l|cc} 
\textbf{Robot type} & \textbf{Avg. \#links} & \textbf{Avg. \#joints} \\ \hline
end effector & 10 & 9 \\ 
robotic arm & 11 & 10 \\ 
humanoid robot & 63 & 62 \\ 
dual arm robot & 33 & 32 \\ 
mobile manipulator & 58 & 58 \\
quadrupedal robot & 19 & 18 \\ 
mobile robot & 13 & 12 \\  \hline
\end{tabular}
\end{table}

\Cref{tab:source_world_flange_count} shows the sources and the number of robots that have links or joints with names containing \emph{world} or \emph{flange}.
These specific words were found to be one of the main differences between the joints and links of the multiply defined robots in \cref{subsec:analysis_duplicate_robots}.
As illustrated, \texttt{ros-industrial} URDF files have a significant number of occurrences  of the word \emph{flange}, implying they may use this convention when developing URDF files.
The word \emph{flange} only appeared in \texttt{ros-industrial} and \texttt{matlab}; which have 14 URDF Bundles with identical files of at least one of the types \textit{.urdf}, \textit{.stl}, \textit{.dae}, or \textit{.dae}.

\begin{table}[!htbp]
\centering
\caption{Number of joints and links with \emph{world} or \emph{flange} in name.}
\label{tab:source_world_flange_count}
\begin{tabular}{l|cc}
\textbf{Source} & \textbf{world} & \textbf{flange} \\ \hline
\texttt{ros-industrial} & 3 & 204  \\ 
\texttt{matlab} & 30 & 14   \\ 
\texttt{robotics-toolbox} & 12 & 0 \\ 
\texttt{drake} & 1 & 0  \\ 
\texttt{oems} & 24 & 0 \\ 
\texttt{random} & 112 & 0  \\ \hline
\end{tabular}
\end{table}

\subsection{Author Contact Information}

In many cases, it may be relevant to know who to contact when working with a URDF model of a robot.
This could be indicated in a comment in the URDF file, by writing the name or email address of the author.
We counted the number of URDF files containing words or symbols related to contact information of authors and found that the word `author' occurred in total in only 13 files, the symbol `@' that can be associated with emails occurred in 16 files, and `.com' occurred in 13 files.
This shows that providing contact information in URDF files is not common.

\subsection{Licensing}
The URDF Bundles in the dataset are protected by different licenses based on the Open Source Initiative~\cite{OpenSourceLicense}.
The four main licenses found in the dataset are shown in \cref{tab:licenses_dataset}.
These licenses are very similar, as they are permissive, meaning developers can use and modify the files, and make their own new versions of them.
The minor differences between these licenses are related to non-endorsement and patenting rights.
When protecting open-source software through licenses, it is natural to choose the most commonly used license within the community, which in this case as shown in \cref{tab:licenses_dataset} is the BSD 3-Clause license.

\newcolumntype{x}[1]{>{\centering\arraybackslash\hspace{0pt}}p{#1}}

\begin{table}
\centering
\caption{Different licenses used in sources of the dataset.}
\label{tab:licenses_dataset}
\begin{tabular}{l|x{1cm}|x{1cm}|x{1cm}|x{1cm}}
\textbf{Source} & \textbf{Apache License v2.0} & \textbf{BSD 3-Clause} & \textbf{BSD 2-Clause} & \textbf{MIT License}  \\ \hline
\texttt{ros-industrial}   & 57 & 45  & 6  & - \\
\texttt{matlab}           & 10 & 36  & 5  & 1   \\
\texttt{robotics-toolbox} & -  & -   & -  & -  \\
\texttt{drake}            & -  & 16  & -  & -   \\
\texttt{oems}             & 13 & 17  & -  & -   \\
\texttt{random}           & 5  & 3   & -  & 6   \\\hline
\textbf{Total}            & 85 & 117 & 11 & 7  \\\hline
\end{tabular}
\end{table}


\section{Construction and Reproducibility} \label{sec:construction}
\noindent This section presents our rationale for how the dataset is constructed.
Additional notes and information can be found in the GitHub repository\footnote{\url{github.com/Daniella1/urdf_dataset_results_material}}. 

\subsection{Construction of the Dataset} \label{subsec:construction}

We have constructed the dataset to represent the general URDF Bundles that can be generated and found on the internet.
We defined categories in an attempt to reduce bias and provide a general representation of URDF Bundles, in order to better analyze their similarities and differences.
Each of the sources must provide URDF Bundles, and fit into one of the categories:
\begin{itemize}
    \item ROS-related sources (\texttt{ros-industrial})
    \item Commercialized tools (\texttt{matlab})
    \item Original Equipment Manufacturers (\texttt{oems})
    \item Common tools used by roboticists (\texttt{robotics-toolbox}, \texttt{drake})
    \item Various repositories that users may find when searching for URDF Bundles (\texttt{random})
\end{itemize}

Although it may be suspected that the quality of the \texttt{random} dataset is lower than any industrial tool's dataset, it is important to include all representations of URDF to understand what a general URDF user may find in their search for URDF Bundles.
The dataset may be biased, by the fact that the \texttt{ros-industrial} dataset contains the larger fraction (34\%) of the URDF files.
This may affect the results when analyzing, for example, folder structures, URDF file generation using xacro, and the types of mesh files.

As the dataset is publicly available, it is possible for others to contribute their URDF Bundles, and perform analyses on the newly added data. 
Instructions on how to add new URDF Bundles or sources are described in the dataset repository.

\subsection{Reproducing Results} \label{subsec:reproducibility}
To reproduce the results in this article, the dataset also contains accompanying Python scripts, located in the GitHub repository in the \emph{scripts/paper\_results} directory\footnote{\url{github.com/Daniella1/urdf_files_dataset}}.
All of the CSV files containing the results presented in this article can be found in the GitHub repository with the dataset results\footnote{\url{github.com/Daniella1/urdf_dataset_results_material}}.

\subsection{URDF Analysis Tool} \label{subsec:analysis_tool}
In addition to the scripts for analyzing the dataset, we are developing a tool for analyzing URDF Bundles.
The tool is publicly available and can be found in the GitHub repository\footnote{\url{github.com/Daniella1/urdf_analyzer}}.
It has been created to operate as a standalone tool, but can also be used together with the dataset.
The tool combines the capabilities of other URDF tools, which are Python-supported and can be used independently of ROS.
The tool can be used to generate the following information:
\begin{description}
    \item[Parser comparison results:] the tool currently supports 6 different URDF parsers, presented in \cref{tab:urdf_parsers_overview}. Results of running all the parsers on the URDF files from the sources of the dataset is shown in \cref{tab:urdf_tool_parsing_results}.

    \item[URDF files parsing results:] this feature can be used to analyze each URDF file specifically with regards to which URDF parsers it failed or succeeded. A URDF file is defined as successfully being parsed, if the result of loading the URDF file with the parser, contains an object and is not `None'. There may be warnings while loading the URDF files, but as long as they can be loaded into an object, we count them as being successfully parsed.

    \item[Comparison of multiply defined robots:] provides information on multiply defined robots, describing the name of the robot, manufacturer, type, sources, and if there are discrepancies in the number of joints or links, mesh types, forward kinematics, and the number of lines in the URDF files.

    \item[Model information:] constructs a table with information on the number and names of the joints and links in each URDF file, and the types of visual and collision meshes used.
\end{description}
\noindent Up to date information about the tool is provided in the repository.

\begin{table}[!htbp]
\centering
\caption{Overview of the URDF parsers currently supported by the tool.}
\label{tab:urdf_parsers_overview}
\begin{tabular}{l|l|l}
\textbf{Parser}  & \textbf{Version} & \textbf{Origin} \\ \hline
\textit{yourdfpy} & 0.0.52   &   PyPi  \\ 
\textit{urdfpy}   & 0.0.22  & PyPi \\ 
\textit{pybullet} & 3.2.5 & PyPi \\
\textit{robotics-toolbox} & 1.0.3  & PyPi \\ 
\textit{MATLAB (R2022b)}  & 9.13  &  PyPi  \\ 
\textit{ROS parser (urdfdom)} & 3.1.0 & conda-forge \\ \hline 
\end{tabular}
\end{table}

The results in \cref{tab:urdf_tool_parsing_results} imply that the so-called Unified Robot Description Format may not be as unified as one would expect.
This may indicate a lack of documentation, forcing the creators of the different URDF parsers to develop parts of the parsing mechanisms using their own understanding of the URDF rules/schema, resulting in a non-unified method for parsing URDF files from different parsers.

\begin{table}[!htbp]
\centering
\caption{Results from running the URDF files from the different sources through the listed parsers. The underlined values are the values where the dataset and parser are from the same organization.}
\label{tab:urdf_tool_parsing_results}
\begin{tabular}{lr|c|c|c|c|c|c}
\textbf{Source} & total & \rot{\textit{yourdfpy}} & \rot{\textit{urdfpy}} & \rot{\textit{pybullet}} & \rot{\textit{robotics-toolbox}} & \rot{\textit{MATLAB}} & \rot{\textit{ROS parser}}\\ \hline
\texttt{ros-industrial}& 108  & 106 & 0 & 90 & 108 & 108 & 108 \\
\texttt{matlab}& \underline{52} & 21 & 0 & 39 & 50 & \underline{52} &  52 \\
\texttt{robotics-toolbox}& \underline{44} & 25 & 0 & 36 & \underline{42}  & 42 & 42 \\
\texttt{drake} & 16  & 8  & 0 & 0 & 14  & 14 & 14 \\ 
\texttt{oems} & 35   & 17  & 1 & 30 & 31 & 31 & 32 \\
\texttt{random}& 67 & 43  & 6 & 57 & 61  & 62  & 63 \\\hline
\textbf{Total}  & 322  & 220 & 7 & 252 &  306 & 309 & 311 \\ \hline
\end{tabular}
\end{table}


\section{Conclusion} \label{sec:ds_conclusion}

\noindent In this article we presented a novel dataset of URDF Bundles accompanied by analyses of the files.
The main highlights of the analyses are:
\begin{itemize}
    \item Only 16/30 of the original robot manufacturers supply a URDF Bundle for at least one of their robots.
    \item 255/270 URDF Bundles were generated using the tool `xacro'.
    \item 11/322 URDF files resulted in errors when parsing them with the official ROS parser, with the most common error being ``XML parsing failed''.
    \item The most commonly used CAD file to visualize URDFs is STL.
    \item The author's contact information is typically not provided in URDF files.
    \item The most commonly used open source software license to protect the URDF files is the BSD 3-Clause license.
\end{itemize}
The results provide us with a better understanding of URDF and the common conventions that can be used when developing a URDF file, such as how to structure the folder and which CAD file types to use.
The tested URDF parsers varied in their performance, showing that the rules/schemas followed by them are inconsistent.

Although, these results present interesting facts about URDF files, there is still more to be studied, for example, validating the visualization of the URDF Bundles was not performed in this study.


\bibliographystyle{IEEEtran}
\bibliography{refs}

\end{document}